\documentclass[conference]{IEEEtran}
\IEEEoverridecommandlockouts
\usepackage{cite}
\usepackage{amsmath,amssymb,amsfonts}
\usepackage{algorithmic}
\usepackage{graphicx}
\usepackage{textcomp}
\usepackage{xcolor}
\def\BibTeX{{\rm B\kern-.05em{\sc i\kern-.025em b}\kern-.08em
    T\kern-.1667em\lower.7ex\hbox{E}\kern-.125emX}}

\usepackage{multirow}
\usepackage{multicol}
\usepackage{todonotes}

\def\method{\textsc{Learnable Word Embedding}}
\def\methodlwe{\textsc{LWE}}
\def\baseline{\textsc{Deeper-lstm-q}}
\def\block{\textsc{block}}

\def\dataset{VQA v\textit{2.0}}
\def\teststd{Test-Standard}
\def\devset{Validation}
\def\testdev{Test-Dev}

\def\softmax{\textsc{softmax}}

\begin{document}

\title{Component Analysis for \\Visual Question Answering Architectures
}


\author{Camila Kolling, J\^onatas Wehrmann, and Rodrigo C. Barros\\
Machine Intelligence and Robotics Research Group\\
School of Technology, Pontifícia Universidade Católica do Rio Grande do Sul\\ 
Av. Ipiranga, 6681, 90619-900, Porto Alegre, RS, Brazil\\
Email: \{camila.kolling,jonatas.wehrmann\}@edu.pucrs.br, rodrigo.barros@pucrs.br
}

\maketitle

\begin{abstract}
Recent research advances in Computer Vision and Natural Language Processing have introduced novel tasks that are paving the way for solving AI-complete problems. One of those tasks is called Visual Question Answering (VQA). This system takes an image and a free-form, open-ended natural-language question about the image, and produce a natural language answer as the output. Such a task has drawn great attention from the scientific community, which generated a plethora of approaches that aim to improve the VQA predictive accuracy. Most of them comprise three major components: (i)~independent representation learning of images and questions; (ii)~feature fusion so the model can use information from both sources to answer visual questions; and (iii)~the generation of the correct answer in natural language. With so many approaches being recently introduced, it became unclear the real contribution of each component for the ultimate performance of the model. The main goal of this paper is to provide a comprehensive analysis regarding the impact of each component in VQA models. Our extensive set of experiments cover both visual and textual elements, as well as the combination of these representations in form of fusion and attention mechanisms. Our major contribution is to identify core components for training VQA models so as to maximize their predictive performance.
\end{abstract}


%
\IEEEpeerreviewmaketitle

\section{Introduction}
%
%
%
%

Recent research advances in Computer Vision (CV) and Natural Language Processing (NLP) introduced several tasks that are quite challenging to be solved, the so-called AI-complete problems. Most of those tasks require systems that understand information from multiple sources, i.e., semantics from visual and textual data, in order to provide some kind of \textit{reasoning}. For instance, image captioning~\cite{donahue2015long, fang2015captions, chen2015mind} presents itself as a hard task to solve, though it is actually challenging to quantitatively evaluate models on that task, and that recent studies~\cite{antol2015vqa} have raised questions on its AI-completeness.

The Visual Question Answering (VQA)~\cite{antol2015vqa} task was introduced as an attempt to solve that issue: to be an actual AI-complete problem whose performance is easy to evaluate. It requires a system that receives as input an image and a free-form, open-ended, natural-language question to produce a natural-language answer as the output \cite{antol2015vqa}. It is a multidisciplinary topic that is gaining popularity by encompassing CV and NLP into a single architecture, what is usually regarded as a multimodal model~\cite{singh2018attention,wehrmann2020aaai,wehrmann2019iccv}. There are many real-world applications for models trained for Visual Question Answering, such as automatic surveillance video queries \cite{tu2014joint} and visually-impaired aiding~\cite{bigham2010vizwiz, lasecki2014increasing}.

Models trained for VQA are required to understand the semantics from images while finding relationships with the asked question. Therefore, those models must present a deep understanding of the image to properly perform inference and produce a reasonable answer to the visual question \cite{xu2016ask}. In addition, it is much easier to evaluate this task since there is a finite set of possible answers for each image-question pair. 

Traditionally, VQA approaches comprise three major steps: (i)~representation learning of the image and the question; (ii)~projection of a single multimodal representation through fusion and attention modules that are capable of leveraging both visual and textual information; and (iii)~the generation of the natural language answer to the question at hand. 
This task often requires sophisticated models that are able to understand a question expressed in text, identify relevant elements of the image, and evaluate how these two inputs correlate.

\begin{figure*}[!htpb]
  \label{fig:basic_net}
  \centering
  \includegraphics[width=0.65\textwidth]{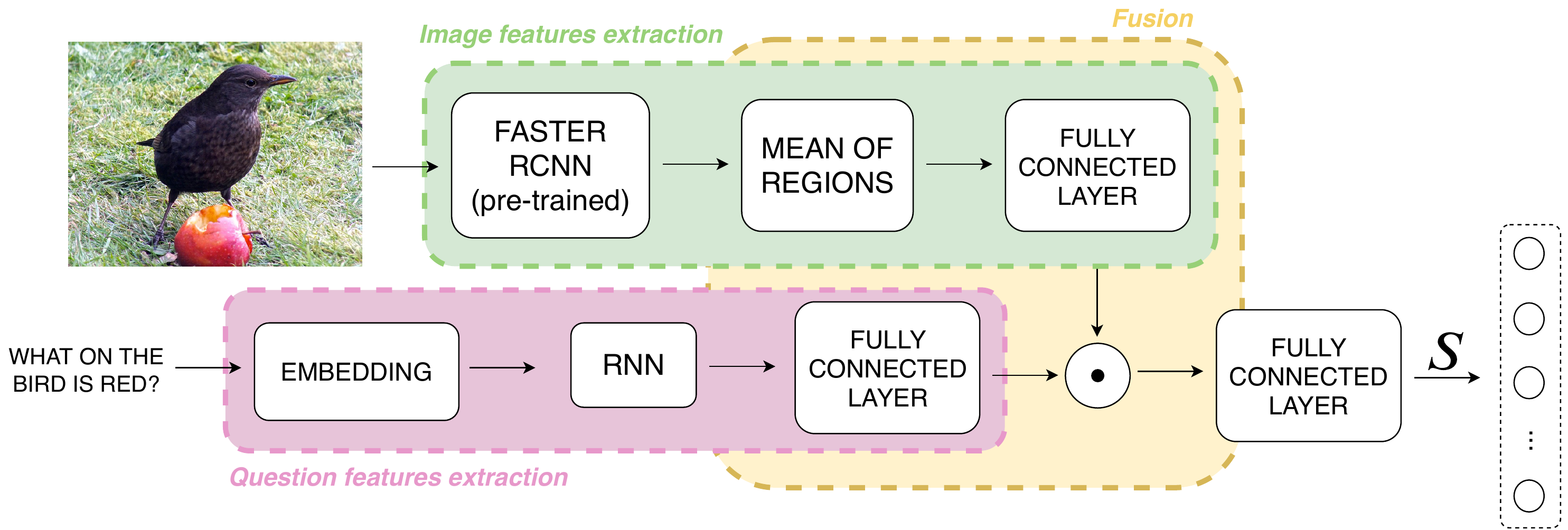}
  \caption{Baseline architecture proposed for the experimental setup.}
\end{figure*}

Given the current interest of the scientific community in VQA, many recent advances try to improve individual components such as the image encoder, the question representation, or the fusion and attention strategies to better leverage both information sources.
With so many approaches currently being introduced at the same time, it becomes unclear the real contribution and importance of each component within the proposed models. Thus, the main goal of this work is to understand the impact of each component on a proposed baseline architecture, which draws inspiration from the pioneer VQA model \cite{antol2015vqa} (Fig.~\ref{fig:basic_net}). Each component within that architecture is then systematically tested, allowing us to understand its impact on the system's final performance through a thorough set of experiments and ablation analysis.

More specifically, we observe the impact of: (i)~pre-trained word embeddings~\cite{pennington2014glove, mikolov2013efficient}, recurrent~\cite{kiros2015skip} and transformer-based sentence encoders~\cite{devlin2018bert} as question representation strategies; (ii)~distinct convolutional neural networks used for visual feature extraction~\cite{ren2015faster, simonyan2014very, he2016deep}; and (iii)~standard fusion strategies, as well as the importance of two main attention mechanisms~\cite{lee2018stacked, anderson2018bottom}. We notice that even using a relatively simple baseline architecture, our best models are competitive to the (maybe overly-complex) state-of-the-art models~\cite{ben2019block,cadene2019murel}. 
Given the experimental nature of this work, we have trained over 130 neural network models, accounting for more than 600 GPU processing hours. We expect our findings to be useful as guidelines for training novel VQA models, and that they serve as a basis for the development of future architectures that seek to maximize predictive performance. 

\section{Related Work}

The task of VAQ has gained attention since Antol et al.~\cite{antol2015vqa} presented a large-scale dataset with  open-ended questions. Many of the developed VQA models employ a very similar architecture \cite{antol2015vqa, fukui2016multimodal, gao2016compact, lu2016hierarchical, malinowski2015ask, mallya2016learning, xiong2016dynamic}: they represent images with features from pre-trained convolutional neural networks; they use word embeddings or recurrent neural networks to represent questions and/or answers; and they combine those features in a classification model over possible answers.

Despite their wide adoption, RNN-based models suffer from their limited representation power \cite{cornia2019m,wehrmann2018cvpr,wehrmann2019fast,wehrmann2017convolutions}. Some recent approaches have investigated the application of the Transformer model \cite{vaswani2017attention} to tasks that incorporate visual and textual knowledge, as image captioning \cite{cornia2019m}. 

Attention-based methods are also being continuously investigated since they enable reasoning by focusing on relevant objects or regions in original input features. They allow models to pay attention on important parts of visual or textual inputs at each step of a task. Visual attention models focus on small regions within an image to extract important features. A number of methods have adopted visual attention to benefit visual question answering \cite{xiong2016dynamic, yang2016stacked, shih2016look}.

Recently, dynamic memory networks \cite{xiong2016dynamic} integrate an attention mechanism with a memory module, and multimodal bilinear pooling \cite{fukui2016multimodal, ben2019block, yu2018beyond} is exploited to expressively combine multimodal features and predict attention over the image. These methods commonly employ visual attention to find critical regions, but textual attention has been rarely incorporated into VQA systems.

While all the aforementioned approaches have exploited those kind of mechanisms, in this paper we study the impact of such choices specifically for the task of VQA, and create a simple yet effective model. Burns et al.~\cite{burns2019language} conducted experiments comparing different word embeddings, language models, and embedding augmentation steps on five multimodal tasks: image-sentence retrieval, image captioning, visual question answering, phrase grounding, and text-to-clip retrieval. While their work focuses on textual experiments, our experiments cover both visual and textual elements, as well as the combination of these representations in form of fusion and attention mechanisms. To the best of our knowledge, this is the first paper that provides a comprehensive analysis on the impact of each major component within a VQA architecture.

\section{Impact of VQA Components}

In this section we first introduce the baseline approach, with default image and text encoders, alongside a pre-defined fusion strategy. That base approach is inspired by the pioneer of Antol~et~al. on VQA~\cite{antol2015vqa}. To understand the importance of each component, we update the base architecture according to each component we are investigating.

In our baseline model we replace the VGG network from~\cite{anderson2018bottom} by a Faster RCNN pre-trained in the Visual Genome dataset~\cite{krishna2017visual}. The default text encoding is given by the last hidden-state of a Bidirectional LSTM network, instead of the concatenation of the last hidden-state and memory cell used in the original work. Fig.~\ref{fig:basic_net} illustrates the proposed baseline architecture, which is subdivided into three major segments: independent feature extraction from (1) images and (2) questions, as well as (3) the fusion mechanism responsible to learn cross-modal features. 

The default text encoder (denoted by the pink rectangle in Fig.~\ref{fig:basic_net}) employed in this work comprises a randomly initialized word-embedding module that takes a tokenized question and returns a continuum vector for each token. Those vectors are used to feed an LSTM network. The last hidden-state is used as the question encoding, which is projected with a linear layer into a $d$-dimensional space so it can be fused along to the visual features. As the default option for the LSTM network, we use a single layer with $2048$ hidden units. Given that this text encoding approach is fully trainable, we hereby name it \method~(\methodlwe). 

For the question encoding, we explore pre-trained and randomly initialized word-embeddings in various settings, including Word2Vec (W2V) \cite{mikolov2013efficient} and GloVe \cite{pennington2014glove}. We also explore the use of hidden-states of Skip-Thoughts Vector \cite{kiros2015skip} and BERT \cite{devlin2018bert} as replacements for word-embeddings and sentence encoding approaches. 

Regarding the visual feature extraction (depicted as the green rectangle in Fig.~\ref{fig:basic_net}), we decided to use the pre-computed features proposed in \cite{anderson2018bottom}. Such an architecture employs a ResNet-152 with a Faster-RCNN \cite{ren2015faster} fine-tuned on the Visual Genome dataset. We opted for this approach due to the fact that using pre-computed features is far more computationally efficient, allowing us to train several models with distinct configurations. Moreover, several recent approaches \cite{ben2019block, cadene2019murel, bai2018deep} employ that same strategy as well, making it easier to provide fair comparison to the state-of-the-art approaches. 
In this study we 
perform experiments with two additional networks widely used for the task at hand, namely VGG-16 \cite{simonyan2014very} and ReSNet-101 \cite{he2016deep}.

Given the multimodal nature of the problem we are dealing with, it is quite challenging to train proper image and question encoders so as to capture relevant semantic information from both of them. Nevertheless, another essential aspect of the architecture is the component that merges them altogether, allowing for the model to generate answers based on both information sources~\cite{duke2018generalized}. The process of multimodal fusion consists itself in a research area with many approaches being recently proposed~\cite{ben2019block, ben2017mutan, fukui2016multimodal, kim2016hadamard}.
The fusion module receives the extracted image and query features, and provides multimodal features that theoretically present information that allows the system to answer to the visual question. There are many fusion strategies that can either assume quite simple forms, such as vector multiplication or concatenation, or be really complex, involving multilayered neural networks, tensor decomposition, and bi-linear pooling, just to name a few. 

Following~\cite{antol2015vqa}, we adopt the element-wise vector multiplication (also referred as Hadamard product) as the default fusion strategy. This approach requires the feature representations to be fused to have the same dimensionality. Therefore, we project them using a fully-connected layer to reduce their dimension from $2048$ to $1024$. After being fused together, the multimodal features are finally passed through a fully-connected layer that provides scores (\textit{logits}) further converted into probabilities via a softmax function ($S$). We want to maximize the probability $P(Y=y|X=x,Q=q)$ of the correct answer $y$ given the image $X$ and the provided question $Q$. Our models are trained to choose within a set comprised by the $3000$ most frequent answers extracted from both training and validation sets of the \dataset~dataset~\cite{goyal2017making}. 

\section{Experimental Setup}



\subsection{Dataset}

For conducting this study we decided to use the \dataset~dataset~\cite{goyal2017making}. It is one of the largest and most frequently used datasets for training and evaluation of models in this task, being the official dataset used in yearly challenges hosted by mainstream computer vision venues~\footnote{VQA Challenge: https://visualqa.org/challenge.html}. This dataset enhances the original one~\cite{antol2015vqa} by alleviating bias problems within the data and increasing the original number of instances.

\dataset~contains over $200,000$ images from MSCOCO \cite{lin2014microsoft}, over 1 million questions and $\approx 11$ million answers. In addition, it has at least two questions per image, which prevents the model from answering the question without considering the input image. 
We follow \dataset~standards and adopt the official provided splits allowing for fair comparison with other approaches. The splits we use are \devset, \testdev, \teststd. 

In this work, results of the ablation experiments are reported on the \devset~set, which is the default option used for this kind of experiment. In some experiments we also report the training set accuracy to verify evidence of overfitting due to excessive model complexity. Training data has a total of $443,757$ questions labeled with $4$ million answers, while the \testdev~has a total of $214,354$ questions. Note that the validation size is about 4-fold larger than ImageNet's, which contains about $50,000$ samples. Therefore, one must keep in mind that even small performance gaps might indicate quite significant results improvement. For instance, 1\% accuracy gains depict $\approx 2,000$ additional instances being correctly classified. We submit the predictions of our best models to the online evaluation servers~\cite{yadav2019evalai} so as to obtain results for the \teststd~split, allowing for a fair comparison to state-of-the-art approaches. 

\subsection{Evaluation Metric}

Free and open-ended questions result in a diverse set of possible answers \cite{antol2015vqa}. For some questions, a simple \textit{yes} or \textit{no} answer may be sufficient. Other questions, however, may require more complex answers. In addition, it is worth noticing that multiple answers may be considered correct, such as \textit{gray} and \textit{light gray}. Therefore, \dataset~provides ten ground-truth answers for each question. These answers were collected from ten different randomly-chosen humans.

The evaluation metric used to measure model performance in the open-ended Visual Question Answering task is a particular kind of accuracy. For each question in the input dataset, the model's most likely response is compared to the ten possible answers provided by humans in the dataset associated with that question \cite{antol2015vqa}, and evaluated according to Equation~\ref{eq:acc}. In this approach, the prediction is considered totally correct only if at least $3$ out of $10$ people provided that same answer.

\begin{figure*}[!htpb]%
\centering
{\includegraphics[width=0.65\textwidth]{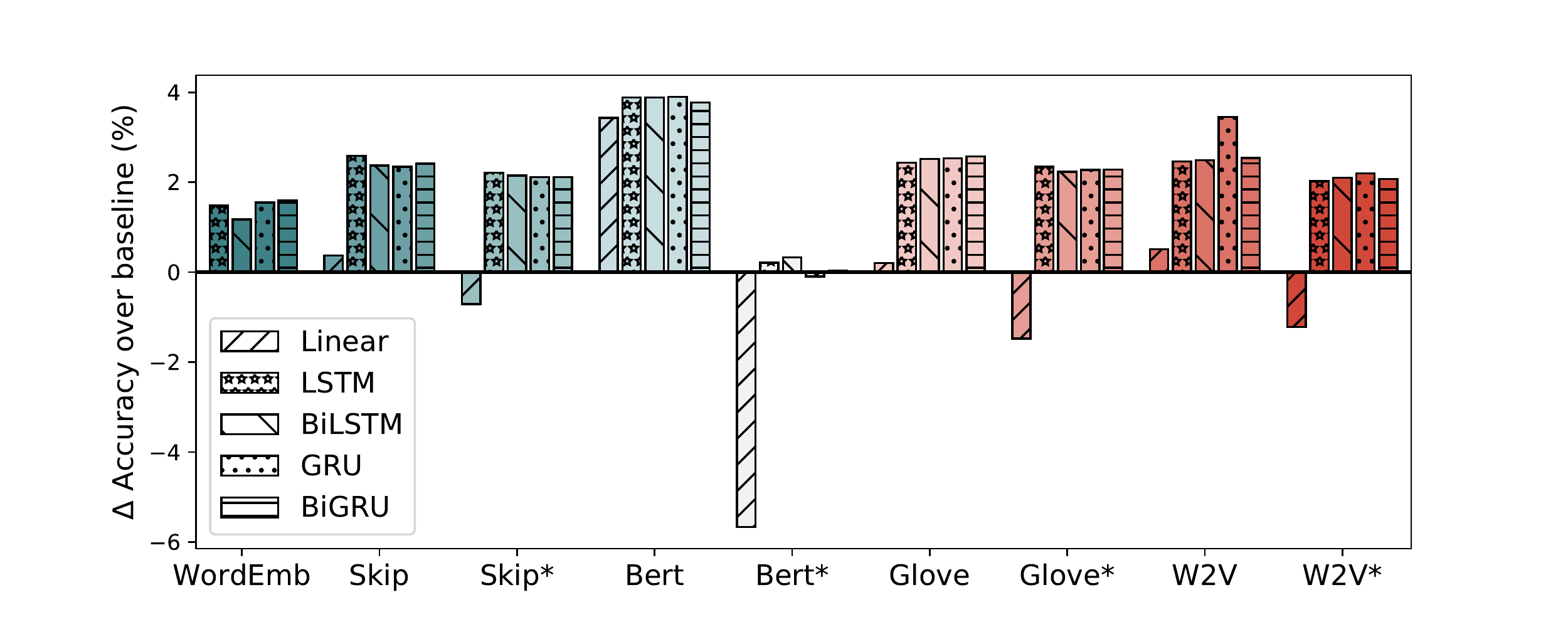}}
\caption{Overall validation accuracy improvement ($\Delta$) over the baseline architecture. Models denoted with \textit{*} present fixed word-embedding representations, i.e., they are not updated via back-propagation.}
\label{fig:delta}
\end{figure*}

\begin{figure}[!htpb]%
\centering
{\includegraphics[width=0.8\columnwidth]{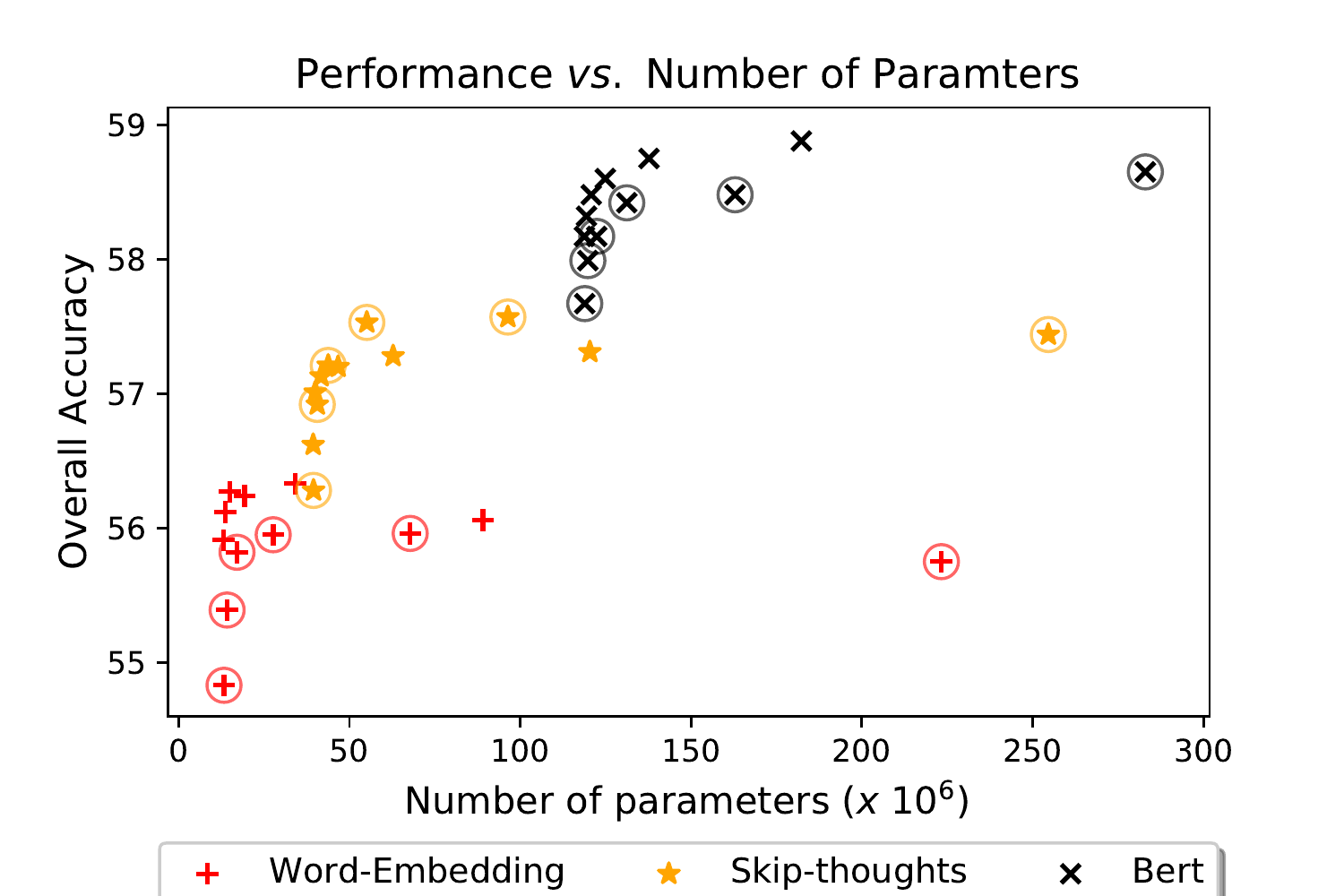}}
\caption{Overall accuracy \textit{vs.} number of parameters trade-off analysis. Circled markers denote two-layered RNNs. Number of parameters increases due to the number of hidden units $H$ within the RNN. In this experiment we vary $H \in \{128, 256, 512, 1024, 2048\}$.}
\label{fig:architecture}
\end{figure}

\begin{equation}
\label{eq:acc}
\centering
\scriptsize
accuracy = \min{\bigg(\frac{\text{\# humans that provided that answer}}{3},1\bigg)}    
\end{equation}

\subsection{Hyper-parameters}

As in \cite{ben2019block} we train our models in a classification-based manner, in which we minimize the cross-entropy loss calculated with an image-question-answer triplet sampled from the training set. We optimize the parameters of all VQA models using Adamax~\cite{kingma2014adam} optimizer with a base learning rate of $7 \times 10^{-4}$, with exception of BERT~\cite{devlin2018bert} in which we apply a 10-fold reduction as suggested in the original paper. We used a learning rate warm-up schedule in which we halve the base learning rate and linearly increase it until the fourth epoch where it reaches twice its base value. It remains the same until the tenth epoch, where we start applying a 25\% decay every two epochs. Gradients are calculated using batch sizes of $64$ instances, and we train all models for 20 epochs.

\section{Experimental Analysis}

In this section we show the experimental analysis for each component in the baseline VQA model. We also provide a summary of our findings regarding the impact of each part. Finally, we train a model with all the components that provide top results and compare it against state-of-the-art approaches.

\subsection{Text Encoder}
\label{sec:txt_enc}

In our first experiment, we analyze the impact of different embeddings for the textual representation of the questions. To this end, we evaluate: (i) the impact of word-embeddings (pre-trained, or trained from scratch); and (ii) the role of the temporal encoding function, i.e., distinct RNN types, as well as pre-trained sentence encoders (e.g., Skip-Thoughts, BERT).

The word-embedding strategies we evaluate are \method~(randomly initialized and trained from scratch), Word2Vec \cite{mikolov2013efficient}, and GloVe \cite{pennington2014glove}. We also use word-level representations from widely used sentence embeddings strategies, namely Skip-Thoughts \cite{kiros2015skip} and BERT \cite{devlin2018bert}. To do so, we use the hidden-states from the Skip-thoughts GRU network, while for BERT we use the activations of the last layer as word-level information. Those vectors feed an RNN that encodes the temporal sequence into a single global vector. 
Different types of RNNs are also investigated for encoding textual representation, including LSTM \cite{hochreiter1997long}, Bidirectional LSTM \cite{schuster1997bidirectional}, GRU \cite{cho2014learning}, and Bidirectional GRU. For bidirectional architectures we concatenate both forward and backward hidden-states so as to aggregate information from both directions. Those approaches are also compared to a linear strategy, where we use a fully-connected layer followed by a global average pooling on the temporal dimension. The linear strategy discards any order information so we can demonstrate the role of the recurrent network as a temporal encoder to improve model performance.



Figure~\ref{fig:delta}~shows the performance variation of different types of word-embeddings, recurrent networks, initialization strategies, and the effect of fine-tuning the textual encoder. 
Clearly, the linear layer is outperformed by any type of recurrent layer. When using Skip-Thoughts the difference reaches $2.22\%$, which accounts for almost $5,000$ instances that the linear model mistakenly labeled. The only case in which the linear approach performed well is when trained with BERT. That is expected since Transformer-based architectures employ several attention layers that present the advantage of achieving the total receptive field size in all layers. While doing so, BERT also encodes temporal information with special positional vectors that allow for learning temporal relations. Hence, it is easier for the model to encode order information within word-level vectors without using recurrent layers.

For the Skip-Thoughts vector model, considering that its original architecture is based on GRUs, we evaluate both the randomly initialized and the pre-trained GRU of the original model, described as [GRU] and [GRU (skip)], respectively. We noticed that both options present virtually the same performance. In fact, GRU trained from scratch performed $0.13\%$ better than its pre-trained version.

Analyzing the results obtained with pre-trained word embeddings, it is clear that GloVe obtained consistently better results than the Word2Vec counterpart. We believe that GloVe vectors perform better given that they capture not only local context statistics as in Word2Vec, but they also incorporate global statistics such as co-occurrence of words.

One can also observe that the use of different RNNs models inflicts minor effects on the results. It might be more advisable to use GRU networks since they halve the number of trainable parameters when compared to the LSTMs, albeit being faster and consistently presenting top results. Note also that the best results for Skip-Thoughts, Word2Vec, and GloVe were all quite similar, without any major variation regarding accuracy.

The best overall result is achieved when using BERT to extract the textual features. BERT versions using either the linear layer or the RNNs outperformed all other pre-trained embeddings and sentence encoders. In addition, the overall training accuracy for BERT models is not so high compared to all other approaches. That might be an indication that BERT models are less prone to overfit training data, and therefore present better generalization ability.   





Results make it clear that when using BERT, one must fine-tune it for achieving top performance. Figure~\ref{fig:delta} shows that it is possible to achieve a $3\%$ to $4\%$ accuracy improvement when updating BERT weights with $1/10$ of the base learning rate. Moreover, Figure \ref{fig:architecture}~shows that the use of a pre-training strategy is helpful, once Skip-thoughts and BERT outperform trainable word-embeddings in most of the evaluated settings. Is also make clear that using a single-layered RNNs provide best results, and are far more efficient in terms of parameters.


\subsection{Image Encoder}
\label{sec:img_enc}

Experiments in this section analyze the visual feature extraction layers. The baseline uses the Faster-RCNN \cite{ren2015faster} network, and we will also experiment with other pre-trained neural networks to encode image information so we can observe their impact on  predictive performance. Additionally to Faster-RCNN, we experiment with two widely used networks for VQA, namely ResNet-101 \cite{he2016deep} and VGG-16 \cite{simonyan2014very}.

\begin{table}[!b]
\centering
\caption{Impact of the network used for visual feature extraction.}
    \begin{tabular}{|c|c|c|c|c|}
        \hline
        \textit{Embedding} & RNN & Network & Training & Validation\\
        \hline
        \multirow{3}{3em}{BERT} & \multirow{3}{2em}{GRU} & Faster & 79.34 & \textbf{58.88} \\
        & & ResNet-101 & 76.14 & 56.09 \\
        & & VGG-16 & 65.59 & 53.49\\
        \hline
    \end{tabular}
    \label{tab:result_visual}
\end{table}

Table~\ref{tab:result_visual} illustrates the result of this experiment. Intuitively, visual features provide a larger impact on model's performance. The accuracy difference between the best and the worst performing approaches is $\approx 5\%$. That difference accounts for roughly $10,000$ validation set instances.
VGG-16 visual features presented the worst accuracy, but that was expected since it is the oldest network used in this study. In addition, it is only sixteen layers deep, and it has been shown that the depth of the network is quite important to hierarchically encode complex structures. Moreover, VGG-16 architecture encodes all the information in a $4096$ dimensional vector that is extracted after the second fully-connected layer at the end. That vector encodes little to none spatial information, which makes it almost impossible for the network to answer questions on the spatial positioning of objects.

ResNet-101 obtained intermediate results. 
It is a much deeper network than VGG-16 and it achieves much better results on ImageNet, which shows the difference of the the learning capacity of both networks. ResNet-101 provides information encoded in $2048$ dimensional vectors, extracted from the global average pooling layer, which also summarizes spatial information into a fixed-sized representation. 

The best result as a visual feature extractor was achieved by the Faster-RCNN fine-tuned on the Visual Genome dataset. Such a network employs a ResNet-152 as backbone for training an RPN-based object detector. In addition, given that it was fine-tuned on the Visual Genome dataset, it allows for the training of robust models suited for general feature extraction. Hence, differently from the previous ResNet and VGG approaches, the Faster-RCNN approach is trained to detect objects, and therefore one can use it to extract features from the most relevant image regions. Each region is encoded as a $2048$ dimensional vector. They contain rich information regarding regions and objects, since object detectors often operate over high-dimensional images, instead of resized ones (e.g., $256 \times 256$) as in typical classification networks. Hence, even after applying global pooling over regions, the network still has access to spatial information because of the pre-extracted regions of interest from each image. 

\begin{table}[!b]
\centering
\caption{Experiment using different fusion strategies.}
    \begin{tabular}{|c|c|c|c|c|c|}
        \hline
        \textit{Embedding} & RNN & Fusion & Training & Validation \\
        \hline
        \multirow{3}{3em}{BERT} & \multirow{3}{2em}{GRU} & Mult & 78.28 & \textbf{58.75} \\ 
        & & Concat & 67.85 & 55.07\\ 
        & & Sum & 68.21 & 54.93\\ 
        \hline
    \end{tabular}
    \label{tab:result_fusion}
\end{table}

\subsection{Fusion strategy}
\label{sec:fus}

In order to analyze the impact that the different fusion methods have on the network performance, three simple fusion mechanisms were analyzed: element-wise multiplication, concatenation, and summation of the textual and visual features.

The choice of the fusion component is essential in VQA architectures, since its output generates multi-modal features used for answering the given visual question. The resulting multi-modal vector is projected into a $3000$-dimensional label space, which provides a probability distribution over each possible answer to the question at hand \cite{duke2018generalized}.

Table~\ref{tab:result_fusion} presents the experimental results with the fusion strategies. The best result is obtained using the element-wise multiplication. Such an approach functions as a filtering strategy that is able to scale down the importance of irrelevant dimensions from the visual-question feature vectors. In other words, vector dimensions with high cross-modal affinity will have their magnitudes increased, differently from the uncorrelated ones that will have their values reduced. Summation does provide the worst results overall, closely followed by the concatenation operator. Moreover, among all the fusion strategies used in this study, multiplication seems to ease the training process as it presents a much higher training set accuracy ($\approx 11\% $ improvement) as well.


\begin{table}[!b]
\centering
\caption{Experiment using different attention mechanisms.}
    \begin{tabular}{|c|c|c|c|c|}
        \hline
        \textit{Embedding} & RNN & Attention & Training & Validation \\
        \hline
        \multirow{5}{2em}{BERT} & \multirow{5}{2em}{GRU} & - & 78.20 & 58.75 \\
        & & Co-Attention & 71.10 & 58.54\\
        & & Co-Attention (L2 norm) & 86.03 & 64.03\\
        & & Top Down & 82.64 & 62.37\\
        & & Top Down ($\sigma=$ReLU) & 87.02 & \textbf{64.12}\\
        \hline
    \end{tabular}
    \label{tab:result_att}
\end{table}

\subsection{Attention Mechanism}
\label{sec:att}

Finally, we analyze the impact of different attention mechanisms, such as Top-Down Attention~\cite{anderson2018bottom} and Co-Attention~\cite{lee2018stacked}. These mechanisms are used to provide distinct image representations according to the asked questions. Attention allows the model to focus on the most relevant visual information required to generate proper answers to the given questions. Hence, it is possible to generate several distinct representations of the same image, which also has a data augmentation effect.

\subsubsection{Top-Down Attention}
\label{sec:topdownatt}

Top-down attention, as the name suggests, uses global features from questions to weight local visual information.
The global textual features $\mathbf{q} \in \mathbb{R}^{2048}$ are selected from the last internal state of the RNN, and the image features $V \in \mathbb{R}^{k \times 2048}$ are extracted from the Faster-RCNN, where $k$ represents the number of regions extracted from the image. In the present work we used $k=36$.
The question features are linearly projected so as to reduce its dimension to $512$, which is the size used in the original paper \cite{anderson2018bottom}. Image features are concatenated with the textual features, generating a matrix $C$ of dimensions $k \times 2560$. Features resulting from that concatenation are first non-linearly projected with a trainable weight matrix $W_1^{2560 \times 512}$ generating a novel multimodal representation for each image region:


\begin{equation}
    \hat{C} = \phi(CW_1)
    \label{eq:fc1}
\end{equation}
Therefore, such a layer learns image-question relations, generating $k \times 512 $ features that are transformed by an activation function $\phi$. Often, $\phi$ is ReLU~\cite{agarap2018deep}, Tanh~\cite{lecun2012efficient}, or Gated Tanh~\cite{xue2018aspect}. The latter employs both the logistic Sigmoid and Tanh, in a gating scheme $\sigma(x) \times \textsc{tanh}(x)$. A second fully-connected layer is employed to summarize the $512$-dimensional vectors into $h$ values per region ($k \times h$). It is usual to use a small value for $h$ such as $\{1, 2\}$. The role of $h$ is to allow the model to produce distinct attention maps, which is useful for understanding complex sentences that require distinct viewpoints. Values produced by this layer are normalized with a \textsc{softmax} function applied on the columns of the matrix, as follows.

\begin{equation}
    A = \textsc{softmax}(\hat{C}W_2)
    \label{eq:fc2}
\end{equation}

It generates an attention mask $A^{k \times h}$ used to weight image regions, producing the image vector $\hat{\mathbf{v}}$, as shown in Equation~\ref{eq:att1}.

\begin{equation}
    \hat{\mathbf{v}_j} = \sum_{i}^{k} V_{i,.} A_{ij}
    \label{eq:att1}
\end{equation}

Note that when $h>1$, the dimensionality of the visual features increases $h$-fold. Hence, $\hat{\mathbf{v}}^{h \times 2048}$, which we reshape to be a $(2048\times h)\times 1$ vector, constitutes the final question-aware image representation.

\subsubsection{Co-Attention}
\label{sec:coatt}

Unlike the Top-Down attention mechanism, Co-Attention is based on the computation of local similarities between all questions words and image regions. It expects two inputs: an image feature matrix $V^{k \times 2048}$, such that each image feature vector encodes an image region out of $k$; and a set of word-level features $Q^{n \times 2048}$. Both $V$ and $Q$ are normalized to have unit $L_2$ norm, so their multiplication $VQ^T$ results in the cosine similarity matrix used as guidance for generating the filtered image features. A context feature matrix $C^{k \times 2048}$ is given by:


\begin{equation}
    C^T = Q^T(QV^T)
\end{equation}

Finally, $C$ is normalized with a $\softmax$~function, and the $k$ regions are summed so as to generate a $1024$-sized vector $\hat{\mathbf{v}}$ to represent relevant visual features $V$ based on question $Q$: 

\begin{equation}
    \hat{\mathbf{v}} = \sum_{i}^{k}\softmax(C)_i
\end{equation}

Table~\ref{tab:result_att}~depicts the results obtained by adding the attention mechanisms to the baseline model. For these experiments we used only element-wise multiplication as fusion strategy, given that it presented the best performance in our previous experiments. We observe that attention is a crucial mechanism for VQA, leading to an $\approx 6\%$ accuracy improvement.

\begin{table*}[!htpb]
\centering
\caption{Comparison of the models on VQA2 \teststd~set. The models were trained on the union of VQA 2.0 trainval split and VisualGenome \cite{krishna2017visual} train split. \textit{All} is the overall OpenEnded accuracy (higher is better). \textit{Yes/No}, \textit{Numbers}, and \textit{Others} are subsets that correspond to answers types. * scores reported from \cite{ben2019block}.}
    \begin{tabular}{|c|c|c|c|c|c|c|c|c|}
        \hline
        \multirow{2}{6em}{Model} & \multicolumn{4}{c|}{VQA2.0 Test-Dev} & \multicolumn{4}{c|}{VQA2.0 Test-Std} \\
        \cline{2-9}
        & All & Yes/No & Num. & Other & All & Yes/No & Num. & Other \\
        \hline
        MCB* \cite{fukui2016multimodal} & - & - & - & - & 62.27 & 78.82 & 38.28 & 53.36 \\
        ReasonNet* \cite{ilievski2017multimodal} & - & - & - & - & 64.61 & 78.86 & 41.98 & 57.39 \\
        Tips\&Tricks* \cite{teney2018tips} & 65.32 & 81.82 & 44.21 & 56.05 & 65.67 & 82.20 & 43.90 & 56.26 \\
        \block*~\cite{ben2019block} & 67.58 & 83.60 & 47.33 & 58.51 & 67.92 & 83.98 & 46.77 & 58.79 \\
        \hline
        BERT-GRU-Faster-TopDown & 67.16 & 84.76 & 44.82 & 57.23 & 67.28 & 84.75 & 44.90 & 57.20\\
        BERT-GRU-Faster-CoAttention &  67.18 & 84.85 & 45.92 & 56.84 & 67.39 & 85.00 & 46.20 & 56.91\\
        \hline
    \end{tabular}
    \label{tab:sota_comp_vqavg}
\end{table*}

\begin{table}[!htpb]
\centering
\caption{Comparison of the models on VQA2 test-dev set. \textit{All} is the overall OpenEnded accuracy (higher is better). \textit{Yes/No}, \textit{Numbers}, and \textit{Others} are subsets that correspond to answers types. * scores reported from \cite{ben2019block}.}
    \begin{tabular}{|c|c|c|c|c|c|c|c|c|}
        \hline
        \multirow{2}{6em}{Model} & \multicolumn{4}{c|}{VQA2.0 Test-Dev} \\
        \cline{2-5}
        & All & Yes/No & Num. & Other \\
        \hline
        \baseline~\cite{antol2015vqa} & 51.95 & 70.42 & 32.28 & 40.64 \\
        MCB* \cite{fukui2016multimodal} & 61.23 & 79.73 & 39.13 & 50.45 \\
        \block*~\cite{ben2019block} & 66.41 & 82.86 & 44.76 & 57.30 \\
        \hline
        BERT-GRU-Faster-CoAttention & 65.84 & 83.66 & 44.36 & 55.50\\
        BERT-GRU-Faster-TopDown & 66.02 & 83.72 & 44.88 & 55.77  \\
        \hline
    \end{tabular}
    \label{tab:sota_comp_vqadev}
\end{table}


The best performing attention approach was Top-Down attention with ReLU activation, followed closely by Co-Attention. We noticed that when using Gated Tanh within Top-Down attention, results degraded 2\%. In addition, experiments show that $L_2$ normalization is quite important in Co-Attention, providing an improvement of almost $6\%$.

\section{Findings Summary}
\label{sec:summary}

The experiments presented in Section~\ref{sec:txt_enc} have shown that the best text encoder approach is fine-tuning a pre-trained BERT model with a GRU network trained from scratch. 

In Section~\ref{sec:img_enc} we performed experiments for analyzing the impact of pre-trained networks to extract visual features, among them Faster-RCNN, ResNet-101, and VGG-16. The best result was using a Faster-RCNN, reaching a $3\%$ improvement in the overall accuracy.

We analyzed different ways to perform multimodal feature fusion in Section~\ref{sec:fus}. In this sense, the fusion mechanism that obtained the best result was the element-wise product. It provides $\approx 3\%$ higher overall accuracy when compared to the other fusion approaches.

Finally, in Section~\ref{sec:att} we have studied two main attention mechanisms and their variations. They aim to provide question-aware image representation by attending to the most important spatial features. The top performing mechanism is the Top-Down attention with the ReLU activation function, which provided an $\approx 6\%$ overall accuracy improvement when compared to the base architecture.

\section{Comparison to state-of-the-art methods}
\label{sec:comparison}

After evaluating individually each component in a typical VQA architecture, our goal in this section is to compare the approach that combines the best performing components into a single model with the current state-of-the-art in VQA. Our comparison involves the following VQA models: \baseline~\cite{antol2015vqa}, MCB~\cite{fukui2016multimodal}, ReasonNet~\cite{ilievski2017multimodal}, Tips\&Tricks~\cite{teney2018tips}, and the recent \block~\cite{ben2019block}.

Tables~\ref{tab:sota_comp_vqavg} and \ref{tab:sota_comp_vqadev} show that our best architecture outperforms all competitors but \block, in both \teststd~(Table~\ref{tab:sota_comp_vqavg}) and \testdev~sets (Table~\ref{tab:sota_comp_vqadev}). Despite~\block~presenting a marginal advantage in accuracy, we have shown in this paper that by carefully analyzing each individual component we are capable of generating a method, without any bells and whistles, that is on par with much more complex methods. 
For instance, \block~and~MCB require 18M and 32M parameters respectively for the fusion scheme alone, while our fusion approach is parameter-free.
Moreover, our model performs far better than \cite{fukui2016multimodal}, \cite{ilievski2017multimodal}, and \cite{teney2018tips}, which are also arguably much more complex methods.

\section{Conclusion}

In this study we observed the actual impact of several components within VQA models. We have shown that transformer-based encoders together with GRU models provide the best performance for question representation. Notably, we demonstrated that using pre-trained text representations provide consistent performance improvements across several hyper-parameter configurations. We have also shown that using an object detector fine-tuned with external data provides large improvements in accuracy. Our experiments have demonstrated that even simple fusion strategies can achieve performance on par with the state-of-the-art. Moreover, we have shown that attention mechanisms are paramount for learning top performing networks, once they allow producing question-aware image representations that are capable of encoding spatial relations. It became clear that Top-Down is the preferred attention method, given its results with ReLU activation. It is is now clear that some configurations used in some architectures (e.g., additional RNN layers) are actually irrelevant and can be removed altogether without harming accuracy.
For future work, we expect to expand this study in two main ways: (i)~cover additional datasets, such as Visual Genome \cite{krishna2017visual}; and (ii)~study in an exhaustive fashion how distinct components interact with each other, instead of observing their impact alone on the classification performance. 

\section*{Acknowledgment}
This study was financed in part by the Coordenação de Aperfeiçoamento de Pessoal de Nivel Superior – Brasil (CAPES) – Finance Code 001. We also would like to thank FAPERGS for funding this research.
We gratefully acknowledge the support of NVIDIA Corporation with the donation of the graphics cards used for this research.





\bibliographystyle{IEEEtran}
\bibliography{main}

\end{document}